\documentclass[runningheads]{llncs}
\usepackage[T1]{fontenc}
\usepackage{xcolor}

\definecolor{darkred}{rgb}{0.9, 0.0, 0.0}
\usepackage{verbatim}
\usepackage[misc]{ifsym}
\usepackage{booktabs}
\usepackage{placeins}
\usepackage{subfigure} 
\usepackage{floatrow}
\usepackage{float}
\usepackage{booktabs}
\usepackage{multirow}
\usepackage[normalem]{ulem}
\useunder{\uline}{\ul}{}
\usepackage{amsmath,amssymb,amsfonts,bm}
\usepackage{graphicx}
\newfloatcommand{capbtabbox}{table}[][\FBwidth]

\begin{document}
\title{All-in-One Medical Image Restoration with Latent Diffusion-Enhanced Vector-Quantized Codebook Prior}

\author{
Haowei Chen\inst{1} \and
Zhiwen Yang\inst{1} \and 
Haotian Hou\inst{1} \and 
Hui Zhang\inst{2} \and 
Bingzheng Wei\inst{3} \and 
Gang Zhou\inst{1} \and
Yan Xu\inst{1}$^{(\textrm{\Letter},\thanks{Corresponding author})}$
} 

% 1{Chen, Haowei}, 2{Yang, Zhiwen}, 3{Hou, Haotian}, 4{Zhang, Hui}, 5{Wei, Bingzheng}, 6{Zhou, Gang}, 7{Xu, Yan}

\authorrunning{Chen et al.}

\institute{
School of Biological Science and Medical Engineering, State Key Laboratory of Software Development Environment, Key Laboratory of Biomechanics and Mechanobiology of Ministry of Education, Beijing Advanced Innovation Center for Biomedical Engineering, Beihang University, Beijing 100191, China
\\
\email{xuyan04@gmail.com} \and 
Department of Biomedical Engineering, Tsinghua University, Beijing 100084, China \and 
ByteDance Inc., Beijing 100098, China
}
    
\maketitle   % typeset the header of the contribution
\begin{abstract}
All-in-one medical image restoration (MedIR) aims to address multiple MedIR tasks using a unified model, concurrently recovering various high-quality (HQ) medical images (e.g., MRI, CT, and PET) from low-quality (LQ) counterparts. However, all-in-one MedIR presents significant challenges due to the heterogeneity across different tasks. Each task involves distinct degradations, leading to diverse information losses in LQ images. Existing methods struggle to handle these diverse information losses associated with different tasks. To address these challenges, we propose a latent diffusion-enhanced vector-quantized codebook prior and develop \textbf{DiffCode}, a novel framework leveraging this prior for all-in-one MedIR. Specifically, to compensate for diverse information losses associated with different tasks, DiffCode constructs a task-adaptive codebook bank to integrate task-specific HQ prior features across tasks, capturing a comprehensive prior. Furthermore, to enhance prior retrieval from the codebook bank, DiffCode introduces a latent diffusion strategy that utilizes the diffusion model's powerful mapping capabilities to iteratively refine the latent feature distribution, estimating more accurate HQ prior features during restoration. With the help of the task-adaptive codebook bank and latent diffusion strategy, DiffCode achieves superior performance in both quantitative metrics and visual quality across three MedIR tasks: MRI super-resolution, CT denoising, and PET synthesis.

% Authors must provide keywords and are not allowed to remove this Keyword section.

\end{abstract}
\section{Introduction}

All-in-one medical image restoration (MedIR) aims to address multiple MedIR tasks using a single, unified model, simultaneously recovering various high-quality (HQ) medical images (e.g., MRI, CT, and PET) from their degraded low-quality (LQ) counterparts. In contrast to current MedIR methods which primarily focus on single-task restoration, such as MRI super-resolution \cite{georgescu2023multimodal,li2024rethinking,zhao2020smore}, CT denoising \cite{chen2023ascon,huang2021gan,ozturk2024denomamba}, and PET synthesis \cite{jang2023spach,luo2022adaptive,zhou2020supervised}, all-in-one MedIR methods offer more versatile solutions for complex clinical scenarios, while improving parameter efficiency and streamlining workflows. 

All-in-one MedIR presents more significant challenges, as it requires managing task heterogeneity across different tasks, which is absent in single-task settings. Specifically, different MedIR tasks are characterized by distinct degradations. These degradations exhibit unique perturbation patterns, resulting in diverse information losses in LQ images \cite{wang2022uformer}. However, existing all-in-one MedIR methods \cite{yang2024all} struggle to adaptively handle these diverse information losses associated with different tasks. Despite the compelling potential of all-in-one MedIR, research in this area remains relatively underexplored.

To address these challenges, the vector-quantized (VQ) codebook prior \cite{lee2022autoregressive,van2017neural} emerges as a promising solution, showcasing two key advantages. First, the VQ codebook prior effectively compensates for LQ images by supplementing rich HQ information. By learning the discrete latent feature representations of undegraded images, a well-trained VQ codebook encodes extensive prior knowledge as HQ prior features within the codebook, with each prior feature representing a fundamental element of HQ image information. Second, the VQ codebook prior is well-suited for handling the task heterogeneity in all-in-one MedIR. Unlike priors in single-task restoration that are typically task-specific, the prior in all-in-one MedIR must be adaptive to different tasks. The VQ codebook prior satisfies this requirement, as VQ codebook priors from different tasks can be seamlessly integrated through a simple codebook concatenation mechanism. This allows us to develop separate codebooks for each task and effortlessly consolidate them into a unified, task-adaptive codebook bank. In this way, the codebook bank integrates the task-specific HQ prior features from each task, providing customized compensation for the diverse information losses associated with different tasks.

However, the effective utilization of this codebook bank requires retrieving accurate HQ prior features that semantically correspond to the LQ degraded features from LQ images during restoration. Such accurate retrieval provides the necessary prior knowledge to consistently compensate for information loss in LQ images. Nevertheless, intrinsic degradation in LQ images corrupts their feature representations, causing a distribution misalignment between the LQ degraded features and the HQ prior features in the latent space. This misalignment prevents the LQ degraded features from accurately retrieving their HQ counterparts, ultimately limiting the full potential of the codebook bank.

To alleviate this misalignment issue, we introduce a latent diffusion strategy \cite{latentdiffusionmodel} that takes advantage of the iterative denoising process of the diffusion model (DM) to refine the latent feature distribution, estimating more appropriate HQ prior features conditioned on the LQ degraded features during restoration. DMs have recently excelled in latent distribution mapping for image restoration tasks \cite{li2024rethinking,luo2023refusion,xia2023diffir}. Unlike direct transformation methods that enforce one-step mapping, DMs gradually reconstruct the complex underlying HQ distribution through controlled noise reduction. This enables the DM to refine a feature distribution that more closely aligns with HQ prior features, thereby facilitating accurate prior retrieval and better exploiting the potential of the codebook bank.

Building on these insights, we propose DiffCode, a novel framework that leverages a latent diffusion-enhanced VQ codebook prior for all-in-one MedIR. DiffCode operates in three stages. Stage I: To compensate for the diverse information losses associated with different tasks, DiffCode constructs a task-adaptive codebook bank for HQ prior encoding, with each codebook in the bank encapsulating task-specific HQ prior features. Stage II: With HQ prior features parameterized in the codebook bank, DiffCode trains a DM in the latent space to optimize prior retrieval. Conditioned on LQ degraded features, DM iteratively refines the feature distribution that better aligns with HQ prior features, providing more accurate HQ features as VQ codebook priors. Stage III: Decoding the HQ prior features retrieved from the codebook bank yields reliable restoration references with enriched HQ information. DiffCode performs the final restoration under the guidance of these references. Furthermore, to mitigate potential task interference \cite{zhao2018modulation}, we adopt a task-aware global routing strategy (TARS) similar to the routing strategy in AMIR \cite{yang2024all}, assigning different tasks to specialized experts throughout the network. The contributions are summarized below.

\begin{figure*}[t]
\label{NA}
    \includegraphics[width=\textwidth]{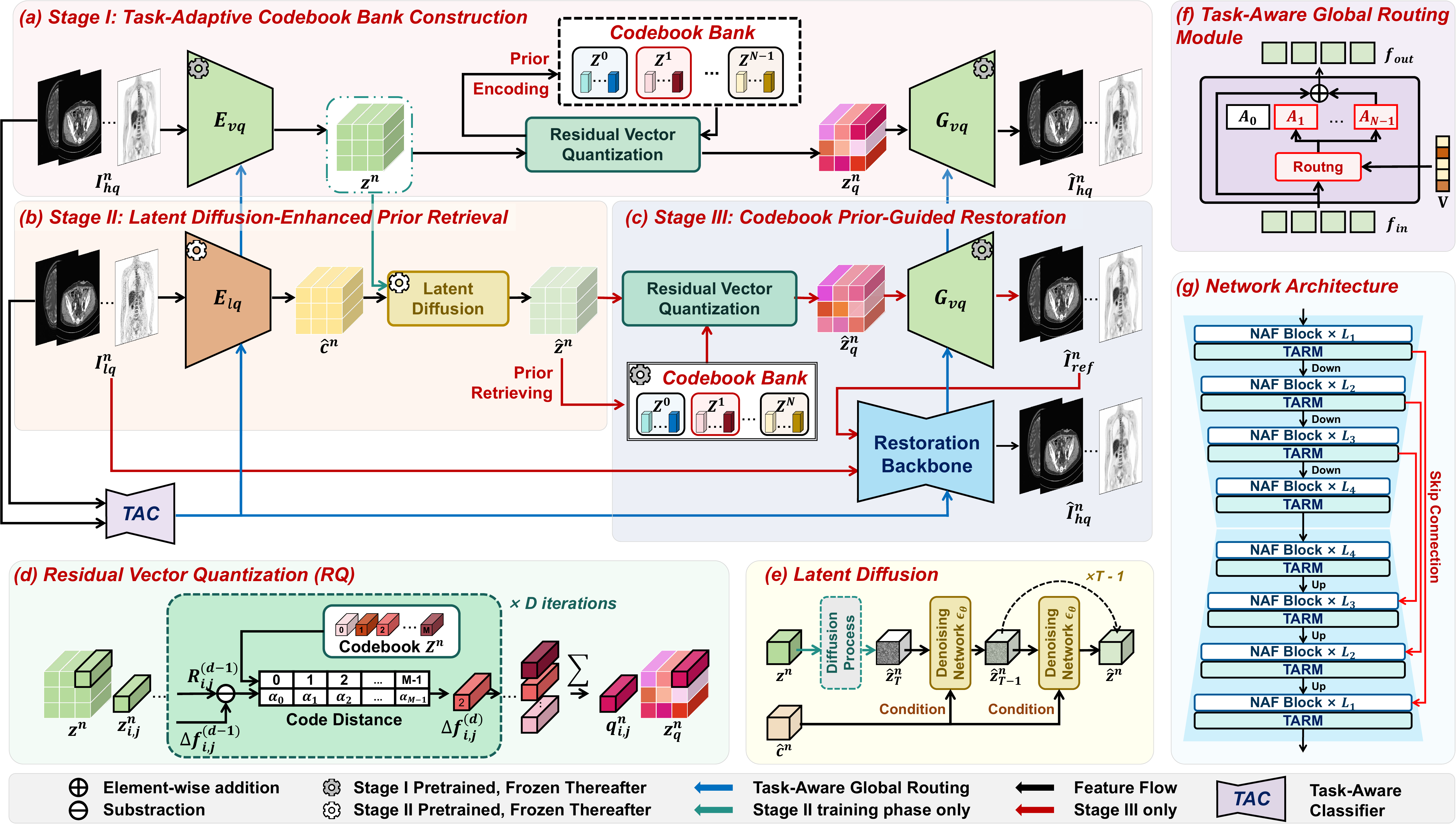}
    \caption{Overview of DiffCode. (a) Stage I constructs a codebook bank to encode HQ priors. (b) Stage II leverages latent diffusion strategy to enhance prior retrieval from the codebook bank. (c) Stage III performs restoration guided by retrieved priors. (d) RQ. (e) DM training. (f) Expert routing in TARM. (g) Network Architecture.} 
    \label{NA}
\end{figure*}

\begin{enumerate}
    \item We are the first to exploit the VQ codebook prior for all-in-one MedIR. By constructing a task-adaptive codebook bank that integrates task-specific prior features from each task, our approach provides customized compensation for the diverse information losses associated with different tasks.
    \item We introduce a latent diffusion strategy that leverages the powerful mapping capability of the DM to enhance prior retrieval from the codebook bank. By iteratively refining the latent feature distributions, the DM estimates more accurate HQ prior features for LQ images during restoration.
    \item Extensive quantitative and qualitative experiments validate the state-of-the-art performance of DiffCode.
\end{enumerate}

\section{Method}
In this section, we first present the three-stage pipeline of our proposed DiffCode in Secs. 2.1–2.3. Furthermore, to mitigate potential task interference, we adopt a task-aware global routing strategy (TARS), which is detailed in Sec. 2.4.

\subsection{Stage I: Task-Adaptive Codebook Bank Construction }
This stage constructs a task-adaptive codebook bank $\{Z^n\}_{n \in [N]}$ for HQ prior encoding, with each codebook $Z^n$ dedicated to the $n^{\text{th}}$ task. For more precise prior encoding, we introduce a residual quantization strategy \cite{lee2022autoregressive}. $\{Z^n\}_{n \in [N]}$ is trained through self-reconstruction learning \cite{van2017neural} on the HQ data. Specifically, HQ images $I^n_{hq}$ from the $n^{\text{th}}$ task are first encoded into latent features $z^n$ by the encoder $E_{vq}$. Then, each element $z^n_{i,j}$ is recursively quantized by the nearest code item $z_m$ in $Z^n$ for $D$ iterations. This operation is known as residual vector quantization (RQ). As shown in Fig. \ref{NA}d, the $d^{\text{th}}$ RQ is formulated as:

\begin{equation}
\mathbf{R}_{i,j}^{(d)} = \mathbf{R}_{i,j}^{(d-1)} - \Delta f_{i,j}^{(d-1)},
\end{equation}
\begin{equation}
\Delta f_{i,j}^{(d)} = \mathop{\operatorname{argmin}}\limits_{z_m \in Z^n} \left\| \mathbf{R}_{i,j}^{(d)} - z_m \right\|^2.
\end{equation}

Where $\Delta f_{i,j}^{(0)} = 0$, $\mathbf{R}_{i,j}^{(d)}$ is the $d^{\text{th}}$ residual and $\mathbf{R}_{i,j}^{(0)} = z^n_{i,j}$. The quantized element $q^n_{i,j} = \sum_{d=1}^{D} \Delta f_{i,j}^{(d)}$. All elements $q^n_{i,j}$ are then combined into quantized features $z^n_q$, which are reconstructed by the decoder $G_{vq}$ for the outputs $\hat{I}^n_{hq}$. 

After Stage I, $\{Z^n\}_{n \in [N]}$ captures a comprehensive VQ codebook prior, with each $Z^n$ encapsulating task-specific HQ prior features for the $n^{\text{th}}$ task.

\subsubsection{Training Strategy.} The total loss $L_{\text{stage1}}$ is defined as

\begin{equation}
L_{\text{stage1}} = \| I_{hq}^n - \hat{I}_{hq}^n \|_1 + \| \mathbf{sg}[z^n] - z_q^n \|_2^2 + \delta \| z^n - \mathbf{sg}[z_q^n] \|_2^2,
\end{equation}

where $\delta=0.25$ followed by the paper \cite{lee2022autoregressive}. The $\| I_{hq}^n - \hat{I}_{hq}^n \|_1$ denotes the reconstruction loss. $\mathbf{sg}[]$ denotes the stop-gradient operator, $\{Z^n\}_{n \in [N]}$ is updated by $\| \mathbf{sg}[z^n] - z_q^n \|_2^2$, and $\| z^n - \mathbf{sg}[z_q^n] \|_2^2$ is the “commitment loss”.

\subsection{Stage II: Latent Diffusion-Enhanced Prior Retrieval}
To retrieve accurate HQ prior features from the codebook bank, we introduce a latent diffusion strategy. As shown in Fig. \ref{NA}b, this strategy employs a DM in the latent space, which comprises two main processes: diffusion and denoising. 

In the diffusion process, the encoder $E_{vq}$ from Stage I is used to extract ground-truth prior features $z^n$ from real HQ images $I^n_{hq}$. These features are then corrupted by noise to produce noisy latent features $\hat{z}^n_T$, as described by:

\begin{equation}
\hat{z}^n_T = \sqrt{\overline{\alpha}_T} z^n + \sqrt{1 - \overline{\alpha}_T} \epsilon,
\end{equation}

where $T$ is the total number of iterations, $\epsilon \sim \mathcal{N}(0,\mathbf{I})$, $\alpha_t=1-\beta_t$, and $\overline{\alpha}_T=\prod_{t=1}^T \alpha_{t}$. The noise variance is controlled by $\beta_{1:T} \in (0,1)$. 

The reverse process begins at the $T^{\text{th}}$ time step and iteratively refines the noisy features $\hat{z}^n_T$ to estimate the potential prior features $\hat{z}^n$. Following the paper \cite{xia2023diffir}, we design a condition encoder $E_{lq}$ to extracts LQ degraded features $\hat{c}^n$ from LQ images $I^n_{lq}$ as conditions. A denoising network $\epsilon_\theta$ subsequently predicts the noise $\epsilon$ conditioned on $\hat{c}^n$ at each step. The reverse step from $\hat{z}^n_t$ to $\hat{z}^n_{t-1}$ is:

\begin{equation}
\hat{z}^n_{t-1} = \frac{1}{\sqrt{\alpha_t}} \left( \hat{z}^n_t - \frac{1 - \alpha_t}{\sqrt{1 - \overline{\alpha}_t}} \epsilon_\theta (\hat{z}^n_t, \hat{c}^n, t) \right) + \sqrt{1 - \alpha_t} \epsilon_t,
\end{equation}

where $\epsilon_t \sim \mathcal{N}(0,\mathbf{I})$. After $T$ iterations of denoising, the DM generates the estimated features $\hat{z}^n$. This iterative refinement process allows $\hat{z}^n$ to better align with the correct prior features $z^n$. The estimated features $\hat{z}^n$ can then be further quantized using $\{Z^n\}_{n \in [N]}$ to retrieve the HQ features $\hat{z}^n_q$ as VQ codebook priors.

During inference, only the reverse process is performed, with the noisy latent features $\hat{z}_T$ initialized from randomly sampled Gaussian noise. 

\subsubsection{Training Strategy.} Latent diffusion requires fewer iterations and smaller sizes compared to traditional DMs \cite{ho2020denoising}. Following the paper \cite{xia2023diffir}, we remove the variance estimation and jointly optimize DM with $E_{lq}$:

\begin{equation}
L_{\text{stage2}} = \| z^n - \hat{z}^n \|_1.
\end{equation}

\subsection{Stage III: Codebook Prior-Guided Restoration}
\textcolor{black}{Stage I and Stage II incur a one-time cost and can provide guidance for training any restoration network.} After retrieving HQ prior features $\hat{z}^n_q$ from the codebook bank $\{Z^n\}_{n \in [N]}$, we employ the decoder $G_{vq}$ to decode them into restoration references $I^n_{ref}$. HQ references $I^n_{ref}$, derived exclusively from the codebook $Z^n$, highlight task-specific HQ details for LQ images $I^n_{lq}$ as guidance. To integrate $I^n_{ref}$, we concatenate $I^n_{ref}$ and $I^n_{lq}$ along the channel dimension and input them into an restoration backbone to recover the final outputs $\hat{I}^n_{hq}$. \textcolor{black}{Notably, this integration step is flexible, and any well-designed fusion method could further enhance performance.}

\subsubsection{Training Strategy.} \textcolor{black}{We use the $L_1$ loss for training, as it is more robust to outliers and better preserves structural details:}

\begin{equation}
L_{\text{stage3}} = \| I^n_{hq} - \hat{I}^n_{hq} \|_1.
\end{equation}

\subsection{Task-aware Global Routing Strategy}
To mitigate task interference \cite{zhao2018modulation}, we adopt a task-aware global routing strategy (TARS) that assigns different tasks to specialized expert networks. \textcolor{black}{In contrast to AMIR \cite{yang2024all}, we employs global routing to prevent expert overuse.} As shown in Fig.  \ref{NA}f and Fig.  \ref{NA}g, we use a mixture-of-expert \cite{shazeer2017sparsely} architecture to construct task-aware global routing modules (TARMs) and insert them into DiffCode. For a given LQ image, a task-aware classifier first extracts its task-aware information to generate the global gating vector $\mathbf{V} \in \mathbb{R}^E$, where $E$ denotes the number of experts in TARM. Then, this vector $\mathbf{V}$ dynamically selects expert networks in TARMs for customized feature processing. TARM is formulated as follows:

\begin{equation}
f_{\text{out}} = \sum_{e \in \mathcal{S}} w_e A_e(f_{\text{in}}), \quad 
w_e = \frac{\exp(\mathbf{V}_e)}{\sum_{s \in \mathcal{S}} \exp(\mathbf{V}_s)}, \quad
\mathcal{S} = \text{TopK}(\mathbf{V}, k).
\end{equation}

Here, $f_{out}$ and $f_{in}$ are the input and output of TARM. $A_e$ represents the $e^{\text{th}}$ expert network, and $k$ controls the number of activated experts in TARM. 

\begin{table}[H]
\caption{Quantitative comparison across three tasks, \textbf{bold} indicates best results.}
\label{Comparison results}
\centering
%\fontsize{8pt}{8pt}\selectfont
\resizebox{\textwidth}{!}{%
\begin{tabular}{@{}c|ccc|ccc|ccc|ccc@{}}
\toprule
\multirow{2}{*}{Method} & \multicolumn{3}{c|}{MRI Super-resolution}       & \multicolumn{3}{c|}{CT Denoising}       & \multicolumn{3}{c|}{PET Synthesis}      & \multicolumn{3}{c}{Average} \\ \cmidrule(l){2-13} 
& PSNR↑  & SSIM↑  & RMSE↓  & PSNR↑  & SSIM↑  & RMSE↓  & PSNR↑ & SSIM↑  & RMSE↓  & PSNR↑  & SSIM↑ & RMSE↓   \\ \midrule
MHCA   & 30.1468   & 0.9149   & 35.5605   & 33.9474   & 0.9062   & 8.2921   & 36.8645   & 0.9420   & 0.0894   & 33.6529  & 0.9210 & 14.6473 \\
Spach Transformer   & 30.6139   & 0.9214   & 33.8595   & 34.0187   & 0.9082   & 8.2354   & 37.1546   & 0.9447   & 0.0865   & 33.9291  & 0.9248 & 14.0605 \\
DenoMamba    & 31.3032   & 0.9307   & 31.4517   & 34.1900   & 0.9087   & 8.0780   & 37.1133   & 0.9454   & 0.0871   & 34.2022  & 0.9283 & 13.2056 \\ \midrule
MPRNet   & 31.0037   & 0.9293   & 32.4893   & 34.0445   & 0.9091   & 8.2138   & 37.2079   & 0.9451   & 0.0860   & 34.0854  & 0.9278 & 13.5964 \\
Restormer    & 31.3467   & 0.9304   & 31.3442   & 34.1855   & 0.9088   & 8.0814   & 37.0901   & 0.9454   & 0.0872   & 34.2074  & 0.9282 & 13.1709 \\
DiffIR   & 31.4398   & 0.9323   & 30.9636   & 34.1933   & 0.9088   & 8.0756   & 37.0980   & 0.9468   & 0.0869   & 34.2437  & 0.9293 & 13.0420 \\ \midrule
AirNet   & 31.1200   & 0.9276   & 32.0352   & 34.2043   & 0.9094   & 8.0626   & 37.0941   & 0.9439   & 0.0870   & 34.1395  & 0.9270 & 13.3949 \\
DRMC     & 29.3003   & 0.8990   & 39.1995   & 33.7899   & 0.9042   & 8.4506   & 36.1101   & 0.9371   & 0.0973   & 33.0668  & 0.9134 & 15.9158 \\
IDR      & 31.2882   & 0.9302   & 31.5034   & 34.1596   & 0.9086   & 8.1073   & 37.1098   & 0.9464   & 0.0871   & 34.1859  & 0.9284 & 13.2326 \\
NDR      & 31.4606   & 0.9334   & 30.9891   & 34.1923   & 0.9089   & 8.0760   & 37.2014   & 0.9464   & 0.0866   & 34.2848  & 0.9296 & 13.0506 \\
AMIR     & 31.6861   & 0.9352   & 30.1821   & 34.2485   & 0.9093   & 8.0267   & 37.1408   & 0.9459   & 0.0872   & 34.3585  & 0.9301 & 12.7653 \\
DiffCode     & \textbf{32.0009} & \textbf{0.9397} & \textbf{29.1957} & \textbf{34.4605} & \textbf{0.9121} & \textbf{7.8381} & \textbf{37.3945} & \textbf{0.9489} & \textbf{0.0845} & \textbf{34.6186}  & \textbf{0.9336} & \textbf{12.3728} \\ \bottomrule
\end{tabular}
}
\end{table}

\begin{figure*}[t]
\label{Comparison}
    \centering
    \includegraphics[width=0.88\textwidth]{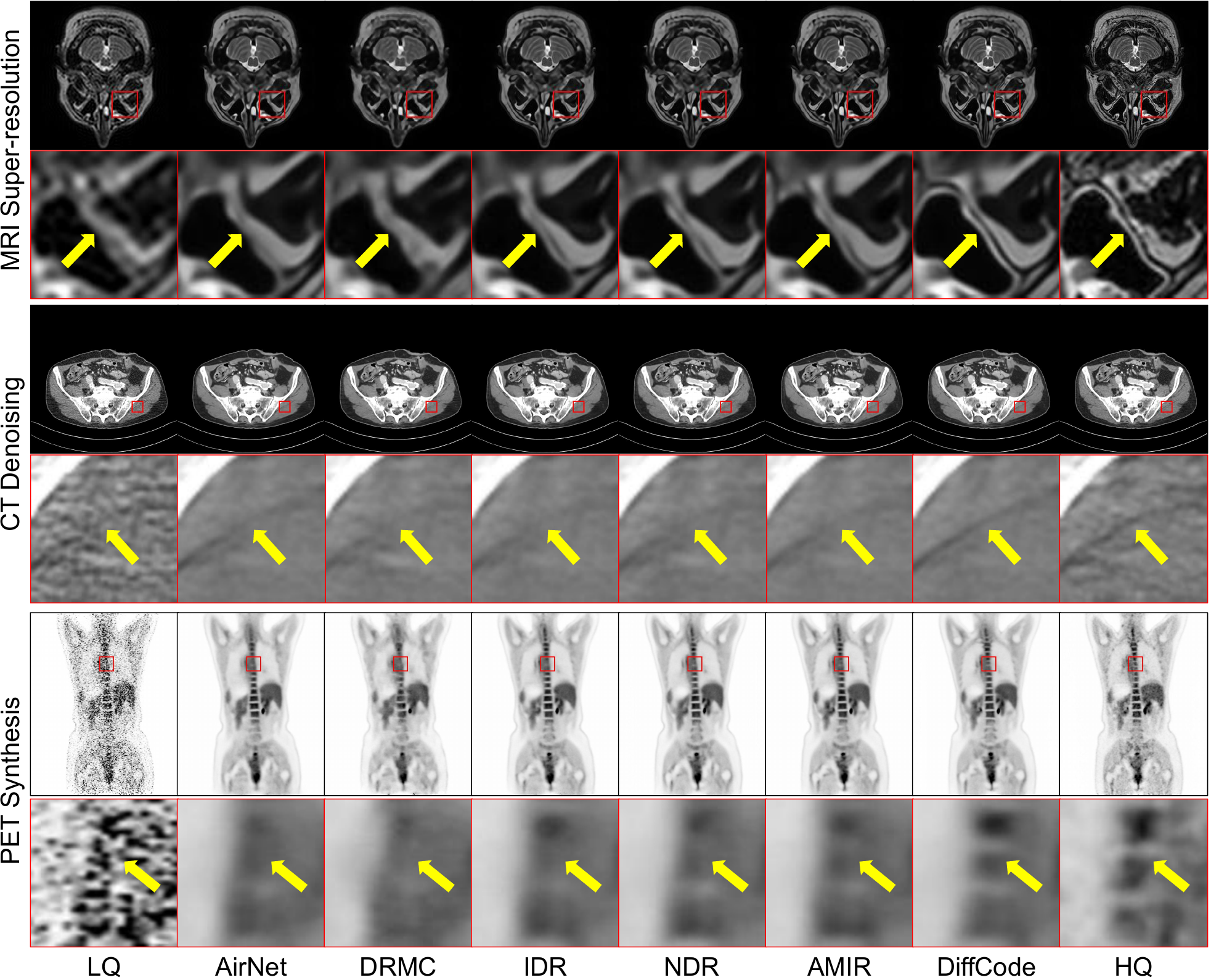}
    \caption{Visual comparison between different methods across three MedIR tasks.} 
    \label{Comparison}
\end{figure*}

\section{Experiment}

\subsection{Dataset and Implementation}

\subsubsection{Dataset.} \textbf{(1) MRI Super-resolution}: The IXI dataset \cite{ixi2022} consists of 578 HQ T2 MRI scans, divided into 405/59/114 for training/validation/testing. For each scan, we extract 100 central 256×256 slices. LQ images are generated using a 4× scaling factor following the paper \cite{zhao2019channel}. \textbf{(2) CT Denoising}: The LDCT dataset \cite{McCollough2020} provides paired standard-dose and quarter-dose CT scans. We select 50 chest scans acquired from a Siemens scanner, divided into 40/5/5 for training/validation/testing, and extract 512×512 slices.  \textbf{(3) PET Synthesis}: A clinical dataset comprising 159 scans is split into 120/10/29 for training/validation/testing. For each scan, we extract 192×400 slices. Following the paper \cite{kim2018penalized}, LQ images are generated by subsampling full scans with a 12× dose reduction, and reconstructed using the standard OSEM method \cite{hudson1994accelerated}.

\subsubsection{Implementation.}
The encoders and decoders utilize [2, 2, 4, 4] NAF blocks \cite{chen2022simple}, with channels [64, 128, 256, 256] at each feature level. \textcolor{black}{NAF block \cite{chen2022simple} is chosen to boost efficiency.} Following the paper \cite{lee2022autoregressive}, each codebook in the codebook bank contains 8192 code items of dimension 256, with the RQ performing 8 iterations. Following the paper \cite{xia2023diffir}, the DM comprises 5 linear layers, with the total time steps set to 8 and $\beta_t$ in Eq.(4) increasing linearly from 0.1 to 0.99. The TARM employs 4 experts followed by AMIR \cite{yang2024all}, with the expert activation number set to 1. Training uses cropped 128×128 patches with a batch size of 8. The model is optimized using the Adam optimizer, starting with a learning rate of 2e-4 and decaying to 1e-6 via cosine annealing.  All experiments are conducted using the PyTorch framework on NVIDIA A100 GPUs. Quantitative performance is evaluated using PSNR, SSIM, and RMSE.

\begin{figure*}[t]
\label{Ablation Visual}
    \includegraphics[width=\textwidth]{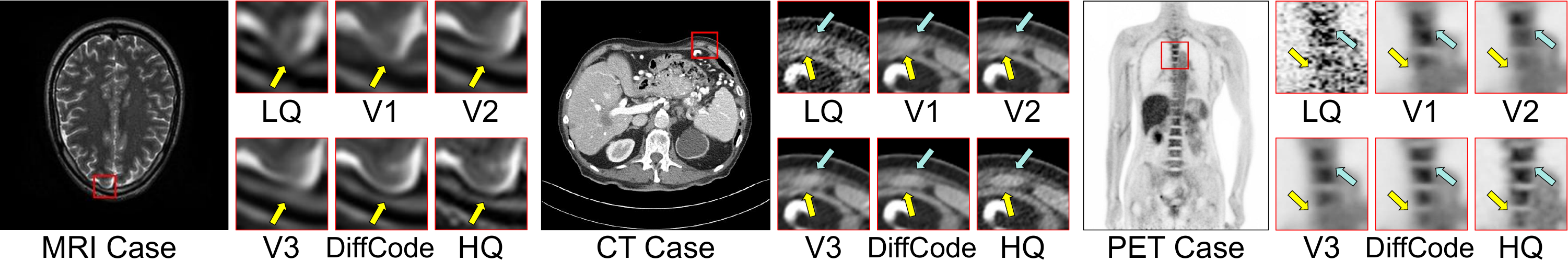}
    \caption{Visual comparison for component analysis of DiffCode across three tasks. The configurations of V1, V2 and V3 are presented in Tab. \ref{Ablation Study} (a).} 
    \label{Ablation Visual}
\end{figure*}

\begin{table}
\caption{\label{Ablation Study}(a) Component analysis of DiffCode, CB refers to the codebook bank, LD refers to latent diffusion. (b) Ablation study on the codebook type and expert activation number in TARS. Performance is the average of three tasks, \textbf{bold} indicates best results. }
\fontsize{8pt}{8pt}\selectfont
\begin{minipage}[b]{0.5\textwidth}
\centering
\resizebox{0.8\textwidth}{!}{%
\begin{tabular}{@{}ccccccc@{}}
\toprule
\multicolumn{7}{c}{(a) Component Analysis} \\ \midrule
\multicolumn{1}{c|}{\multirow{2}{*}{Method}} & \multicolumn{3}{c|}{Configuration} & \multicolumn{3}{c}{Performance} \\ \cmidrule(l){2-7} 
\multicolumn{1}{c|}{} & CB & LD & \multicolumn{1}{c|}{TARS} & PSNR↑ & SSIM↑ & RMSE↓ \\ \cmidrule(l){1-7} 
\multicolumn{1}{c|}{V1} &  &  & \multicolumn{1}{c|}{} & 34.2057 & 0.9285 & 13.2157 \\
\multicolumn{1}{c|}{V2} &  &  & \multicolumn{1}{c|}{\checkmark} & 34.4079 & 0.9305 & 13.1176 \\
\multicolumn{1}{c|}{V3} & \checkmark &  & \multicolumn{1}{c|}{\checkmark} & 34.5273 & 0.9323 & 12.8168 \\
\multicolumn{1}{c|}{DiffCode} & \checkmark & \checkmark & \multicolumn{1}{c|}{\checkmark} & {\textbf{34.6186}} & {\textbf{0.9336}} & {\textbf{12.3728}} \\ \bottomrule
\end{tabular}
}
\end{minipage}% 
\begin{minipage}[b]{0.5\textwidth}
\centering
\resizebox{0.72\textwidth}{!}{%
\begin{tabular}{@{}cccccc@{}}
\toprule
\multicolumn{6}{c}{(b) Ablation on Codebook and TARS} \\ \midrule
\multicolumn{1}{c|}{\multirow{2}{*}{TARS}} & \multicolumn{2}{c|}{Codebook} & \multicolumn{3}{c}{Performance} \\ \cmidrule(l){2-6} 
\multicolumn{1}{l|}{} & Single & \multicolumn{1}{c|}{Bank} & PSNR↑ & SSIM↑ & RMSE↓ \\ \cmidrule(l){1-6} 
\multicolumn{1}{c|}{3} &  & \multicolumn{1}{c|}{\checkmark} & 34.5425 & 0.9329 & 12.6693 \\
\multicolumn{1}{c|}{2} &  & \multicolumn{1}{c|}{\checkmark} & 34.5687 & 0.9330 & 12.7238 \\
\multicolumn{1}{c|}{1} & \checkmark & \multicolumn{1}{c|}{} & 34.5666 & 0.9332 & 12.6825 \\
\multicolumn{1}{c|}{1} &  & \multicolumn{1}{c|}{\checkmark} & {\textbf{34.6186}} & {\textbf{0.9336}} & {\textbf{12.3728}} \\ \bottomrule
\end{tabular}
}
\end{minipage}
\end{table}

\subsection{Comparison and Ablation Study}
\subsubsection{Comparison.}
We compare DiffCode against three task-specific MedIR methods: MHCA \cite{georgescu2023multimodal}, Spach Transformer \cite{jang2023spach}, DenoMamba \cite{ozturk2024denomamba}; three general image restoration methods: MPRNet \cite{zamir2021multi}, Restormer \cite{zamir2022restormer}, DiffIR \cite{xia2023diffir}; and five all-in-one image restoration methods: AirNet \cite{li2022all}, DRMC \cite{yang2023drmc}, IDR \cite{zhang2023ingredient}, NDR \cite{yao2024neural}, AMIR \cite{yang2024all}. All competing methods are trained in an all-in-one setting. As observed in Tab. \ref{Comparison results} and Fig. \ref{Comparison}, DiffCode outperforms all competing methods in both quantitative metrics and visual quality, demonstrating superior performance.

\subsubsection{Ablation Study.} The component analysis in Tab. \ref{Ablation Study}a and Fig. \ref{Ablation Visual} confirms the necessity of three core components: the codebook bank, the latent diffusion strategy, and TARS. Comparing V2 to V1, TARS assigns tasks to specialized experts, mitigating task interference for better performance. Integrating the codebook bank into V3 provides LQ images with HQ priors, further boosting restoration quality. Finally, DiffCode surpasses V3 by introducing a latent diffusion strategy that enhances prior retrieval during restoration. Moreover, the ablation study in Tab. \ref{Ablation Study}b highlights two critical insights: First, the codebook bank outperforms a single-codebook design by encoding task-specific priors. Second, optimal performance emerges when TARS activates individual experts for each task. This is likely because medical images from different tasks exhibit distinct statistical distributions, which may cause task interference when the same expert is shared.

\section{Conclusion}
In this paper, we propose a latent diffusion-enhanced VQ codebook prior and develop DiffCode, a novel framework that leverages this prior for all-in-one MedIR. By constructing a task-adaptive codebook bank that integrates task-specific HQ prior features across different tasks, DiffCode provides customized compensation for the diverse information losses associated with each task. Moreover, DiffCode introduces a latent diffusion strategy to enhance prior retrieval during restoration, iteratively refining the latent feature distribution to estimate more accurate HQ prior features for LQ images. Experimental results demonstrate the superiority of DiffCode across three MedIR tasks. In the future, we will explore the effectiveness of the proposed DiffCode as more MedIR tasks are involved.

\begin{credits}
\subsubsection{\ackname} 
This work is supported by the National Natural Science Foundation in China under Grant 62371016, U23B2063, 62022010, and 62176267, the Bejing Natural Science Foundation Haidian District Joint Fund in China under Grant L222032, the Beijing hope run special fund of cancer foundation of China under Grant LC2018L02, the Fundamental Research Funds for the Central University of China from the State Key Laboratory of Software Development Environment in Beihang University in China, the 111 Proiect in China under Grant B13003, the SinoUnion Healthcare Inc. under the eHealth program, the high performance computing (HPC) resources at Beihang University.

\subsubsection{\discintname}
We have no conflicts of interest to disclose.
\end{credits}

% \usepackage{booktabs}
% \usepackage{multirow}
% \usepackage[normalem]{ulem}

% \usepackage{booktabs}
% \usepackage{multirow}
% \usepackage{multirow}
% \usepackage{multirow}

\bibliographystyle{splncs04}
\bibliography{Paper-1126} 

\begin{thebibliography}{10}
\providecommand{\url}[1]{\texttt{#1}}
\providecommand{\urlprefix}{URL }
\providecommand{\doi}[1]{https://doi.org/#1}

\bibitem{ixi2022}
Ixi dataset. \url{https://brain-development.org/ixi-dataset/}

\bibitem{chen2022simple}
Chen, L., Chu, X., Zhang, X., Sun, J.: Simple baselines for image restoration. In: European conference on computer vision. pp. 17--33. Springer (2022)

\bibitem{chen2023ascon}
Chen, Z., Gao, Q., Zhang, Y., Shan, H.: Ascon: Anatomy-aware supervised contrastive learning framework for low-dose ct denoising. In: International Conference on Medical Image Computing and Computer-Assisted Intervention. pp. 355--365. Springer (2023)

\bibitem{georgescu2023multimodal}
Georgescu, M.I., Ionescu, R.T., Miron, A.I., Savencu, O., Ristea, N.C., Verga, N., Khan, F.S.: Multimodal multi-head convolutional attention with various kernel sizes for medical image super-resolution. In: Proceedings of the IEEE/CVF winter conference on applications of computer vision. pp. 2195--2205 (2023)

\bibitem{ho2020denoising}
Ho, J., Jain, A., Abbeel, P.: Denoising diffusion probabilistic models. Advances in neural information processing systems  \textbf{33},  6840--6851 (2020)

\bibitem{huang2021gan}
Huang, Z., Zhang, J., Zhang, Y., Shan, H.: Du-gan: Generative adversarial networks with dual-domain u-net-based discriminators for low-dose ct denoising. IEEE Transactions on Instrumentation and Measurement  \textbf{71},  1--12 (2021)

\bibitem{hudson1994accelerated}
Hudson, H.M., Larkin, R.S.: Accelerated image reconstruction using ordered subsets of projection data. IEEE transactions on medical imaging  \textbf{13}(4),  601--609 (1994)

\bibitem{jang2023spach}
Jang, S.I., Pan, T., Li, Y., Heidari, P., Chen, J., Li, Q., Gong, K.: Spach transformer: Spatial and channel-wise transformer based on local and global self-attentions for pet image denoising. IEEE transactions on medical imaging  (2023)

\bibitem{kim2018penalized}
Kim, K., Wu, D., Gong, K., Dutta, J., Kim, J.H., Son, Y.D., Kim, H.K., El~Fakhri, G., Li, Q.: Penalized pet reconstruction using deep learning prior and local linear fitting. IEEE transactions on medical imaging  \textbf{37}(6),  1478--1487 (2018)

\bibitem{lee2022autoregressive}
Lee, D., Kim, C., Kim, S., Cho, M., Han, W.S.: Autoregressive image generation using residual quantization. In: Proceedings of the IEEE/CVF Conference on Computer Vision and Pattern Recognition. pp. 11523--11532 (2022)

\bibitem{li2022all}
Li, B., Liu, X., Hu, P., Wu, Z., Lv, J., Peng, X.: All-in-one image restoration for unknown corruption. In: Proceedings of the IEEE/CVF conference on computer vision and pattern recognition. pp. 17452--17462 (2022)

\bibitem{li2024rethinking}
Li, G., Rao, C., Mo, J., Zhang, Z., Xing, W., Zhao, L.: Rethinking diffusion model for multi-contrast mri super-resolution. In: Proceedings of the IEEE/CVF Conference on Computer Vision and Pattern Recognition. pp. 11365--11374 (2024)

\bibitem{luo2022adaptive}
Luo, Y., Zhou, L., Zhan, B., Fei, Y., Zhou, J., Wang, Y., Shen, D.: Adaptive rectification based adversarial network with spectrum constraint for high-quality pet image synthesis. Medical Image Analysis  \textbf{77},  102335 (2022)

\bibitem{luo2023refusion}
Luo, Z., Gustafsson, F.K., Zhao, Z., Sj{\"o}lund, J., Sch{\"o}n, T.B.: Refusion: Enabling large-size realistic image restoration with latent-space diffusion models. In: Proceedings of the IEEE/CVF conference on computer vision and pattern recognition. pp. 1680--1691 (2023)

\bibitem{McCollough2020}
McCollough, C., Chen, B., Holmes~III, D.R., Duan, X., Yu, Z., Yu, L., Leng, S., Fletcher, J.: Low dose ct image and projection data (ldct-and-projection-data) (version 6) [data set] (2020). \doi{10.7937/9NPB-2637}

\bibitem{ozturk2024denomamba}
{\"O}zt{\"u}rk, {\c{S}}., Duran, O.C., {\c{C}}ukur, T.: Denomamba: A fused state-space model for low-dose ct denoising. arXiv preprint arXiv:2409.13094  (2024)

\bibitem{latentdiffusionmodel}
Rombach, R., Blattmann, A., Lorenz, D., Esser, P., Ommer, B.: High-resolution image synthesis with latent diffusion models. In: Proceedings of the IEEE/CVF conference on computer vision and pattern recognition. pp. 10684--10695 (2022)

\bibitem{shazeer2017sparsely}
Shazeer, N., Mirhoseini, A., Maziarz, K., Davis, A., Le, Q., Hinton, G., Dean, J.: The sparsely-gated mixture-of-experts layer. Outrageously large neural networks  (2017)

\bibitem{van2017neural}
Van Den~Oord, A., Vinyals, O., et~al.: Neural discrete representation learning. Advances in neural information processing systems  \textbf{30} (2017)

\bibitem{wang2022uformer}
Wang, Z., Cun, X., Bao, J., Zhou, W., Liu, J., Li, H.: Uformer: A general u-shaped transformer for image restoration. In: Proceedings of the IEEE/CVF conference on computer vision and pattern recognition. pp. 17683--17693 (2022)

\bibitem{xia2023diffir}
Xia, B., Zhang, Y., Wang, S., Wang, Y., Wu, X., Tian, Y., Yang, W., Van~Gool, L.: Diffir: Efficient diffusion model for image restoration. In: Proceedings of the IEEE/CVF International Conference on Computer Vision. pp. 13095--13105 (2023)

\bibitem{yang2024all}
Yang, Z., Chen, H., Qian, Z., Yi, Y., Zhang, H., Zhao, D., Wei, B., Xu, Y.: All-in-one medical image restoration via task-adaptive routing. In: International Conference on Medical Image Computing and Computer-Assisted Intervention. pp. 67--77. Springer (2024)

\bibitem{yang2023drmc}
Yang, Z., Zhou, Y., Zhang, H., Wei, B., Fan, Y., Xu, Y.: Drmc: A generalist model with dynamic routing for multi-center pet image synthesis. In: International Conference on Medical Image Computing and Computer-Assisted Intervention. pp. 36--46. Springer (2023)

\bibitem{yao2024neural}
Yao, M., Xu, R., Guan, Y., Huang, J., Xiong, Z.: Neural degradation representation learning for all-in-one image restoration. IEEE Transactions on Image Processing  (2024)

\bibitem{zamir2022restormer}
Zamir, S.W., Arora, A., Khan, S., Hayat, M., Khan, F.S., Yang, M.H.: Restormer: Efficient transformer for high-resolution image restoration. In: Proceedings of the IEEE/CVF conference on computer vision and pattern recognition. pp. 5728--5739 (2022)

\bibitem{zamir2021multi}
Zamir, S.W., Arora, A., Khan, S., Hayat, M., Khan, F.S., Yang, M.H., Shao, L.: Multi-stage progressive image restoration. In: Proceedings of the IEEE/CVF conference on computer vision and pattern recognition. pp. 14821--14831 (2021)

\bibitem{zhang2023ingredient}
Zhang, J., Huang, J., Yao, M., Yang, Z., Yu, H., Zhou, M., Zhao, F.: Ingredient-oriented multi-degradation learning for image restoration. In: Proceedings of the IEEE/CVF Conference on Computer Vision and Pattern Recognition. pp. 5825--5835 (2023)

\bibitem{zhao2020smore}
Zhao, C., Dewey, B.E., Pham, D.L., Calabresi, P.A., Reich, D.S., Prince, J.L.: Smore: a self-supervised anti-aliasing and super-resolution algorithm for mri using deep learning. IEEE transactions on medical imaging  \textbf{40}(3),  805--817 (2020)

\bibitem{zhao2018modulation}
Zhao, X., Li, H., Shen, X., Liang, X., Wu, Y.: A modulation module for multi-task learning with applications in image retrieval. In: Proceedings of the European Conference on Computer Vision (ECCV). pp. 401--416 (2018)

\bibitem{zhao2019channel}
Zhao, X., Zhang, Y., Zhang, T., Zou, X.: Channel splitting network for single mr image super-resolution. IEEE transactions on image processing  \textbf{28}(11),  5649--5662 (2019)

\bibitem{zhou2020supervised}
Zhou, L., Schaefferkoetter, J.D., Tham, I.W., Huang, G., Yan, J.: Supervised learning with cyclegan for low-dose fdg pet image denoising. Medical image analysis  \textbf{65},  101770 (2020)

\end{thebibliography}
\end{document}